\documentclass{article}
\usepackage[utf8]{inputenc}
\usepackage{spconf,graphicx}
\usepackage{comment}
\usepackage{amsmath,amssymb}
\usepackage{color}
\usepackage{url}
\usepackage{hyperref}
\usepackage[export]{adjustbox}
\usepackage{subcaption}
\usepackage{tcolorbox}
\newcommand{\todom}[1]{} 
\newcommand{\C}{\mathbb{C}}
\newcommand{\R}{\mathbb{R}}
\newcommand{\E}{\mathbb{E}}

\newcommand{\real}{\mathcal{R}}
\newcommand{\imag}{\mathcal{I}}
\newcommand{\rCov}{\text{Cov}} 
\newcommand{\Cov}{\text{Cov}_{\C}}
\newcommand{\Var}{\text{Var}}
\DeclareMathOperator{\sgn}{sgn}
\definecolor{orange}{rgb}{0.470588235,0.470588235,0}
\definecolor{pink}{rgb}{0.294117647,0,0.294117647}
\definecolor{myblue}{rgb}{0, 0, 0.647058824}
\definecolor{mygreen}{rgb}{0, 0.823529412, 0}
\definecolor{myred}{rgb}{1, 0, 0}

\title{Building Blocks for a Complex-Valued Transformer Architecture}
\name{Florian Eilers\sthanks{Member of the CiM IMPRS Graduate School Münster}, Xiaoyi Jiang\sthanks{This work was partly supported by the Deutsche Forschungsgemeinschaft
(DFG) – CRC 1450 – 431460824 and European Union’s Horizon 2020
under the Marie Sklodowska-Curie grant agreement No 778602 Ultracept.
}}
\address{Faculty of Mathematics and Computer Science, University of Münster,
Münster, Germany}
\begin{document}
\ninept
\maketitle
\begin{abstract}
Most deep learning pipelines are built on real-valued operations to deal with real-valued inputs such as images, speech or music signals.
However, a lot of applications naturally make use of complex-valued signals or images, such as MRI or remote sensing.
Additionally the Fourier transform of signals is complex-valued and has numerous applications.
We aim to make deep learning directly applicable to these complex-valued signals without using projections into $\R^2$.
Thus we add to the recent developments of complex-valued neural networks by presenting building blocks to transfer the transformer architecture to the complex domain.
We present multiple versions of a complex-valued Scaled Dot-Product Attention mechanism as well as a complex-valued layer normalization.
We test on a classification and a sequence generation task on the MusicNet dataset and show improved robustness to overfitting while maintaining on-par performance when compared to the real-valued transformer architecture.

\end{abstract}
\section{Introduction}

In recent years, many applications have benefited from the fast development and high quality results of deep learning methods.
Most of these methods focus on real-valued pipelines for applications with real-valued signals, such as natural images or encodings of natural language processing. 
There is however a great amount of applications that naturally deal with complex-valued signals, such as MRI images \cite{cole2021analysis, vasudeva2022compressed} or remote sensing \cite{li2020sscv} and the Fourier transform of real-valued signals \cite{choi2018phase, yang2020complex} or images \cite{yang2020fda, xu2021fourier} and it has been shown that fully complex-valued architectures often (but not always \cite{yin2020phasen}) deliver superior performance when dealing with complex-valued signals.
The complex numbers come with an intrinsic algebraic structure that can not be captured by the simple isomorphism of $\C \sim \R^2$, especially because there is no natural way to define multiplication in $\R^2$, which, however, is an important part of many deep learning building blocks.
\cite{Trabel18deepcom} has provided a lot of those building blocks, such as complex-valued convolution, batch normalization and initialization.
These building blocks are of great help for a large amount of current architectures, especially in image and signal processing.
In many fields, architectures building on the idea of attention mechanisms have successfully been applied.
Especially the immense success of the transformer architecture \cite{vaswani2017attention} has shown that attention based architectures can be superior and have since become standard in many applications.
We seek to provide a solid generalization of the building blocks of the transformer architecture in the complex domain and show experimental evidence that it improves robustness to overfitting while maintaining on-par performance when compared to the real-valued transformer architecture.

Our key contributions are:
%
\textbf{1)} Newly developed building blocks consisting of:
a. derivation of a complex-valued attention mechanism, generalizing the Scaled Dot-Product attention \cite{vaswani2017attention};
b. introduction of complex-valued layer normalization.
\textbf{2)} Adaptation of building blocks from existing complex-valued neural networks for the transformer architecture.
\textbf{3)} Demonstration of improved robustness to overfitting while maintaining on-par results compared to the real-valued model.

The combination of the first two contributions provide the foundation for a mathematically rigorous complex-valued transformer architecture.
The source code for the full architecture and all experiments is available as a Pytorch module
\footnote{\url{https://zivgitlab.uni-muenster.de/ag-pria/cv-transformer}}.

\section{Related work}
Complex-valued neural networks have been researched for a long time \cite{kataoka1998music, kuroe2002energy}.
An early standard book and foundation for much research to come is the work by Hirose \cite{hirose2003complex}. 
Recently, an increasing number of works in complex-valued neural networks have been published \cite{bassey2021survey}, driven by the interest in applications, which naturally deal with complex-valued signals: remote sensing \cite{zhang2021end, ren2021polsar}, MRI processing \cite{virtue2017better, cole2021analysis} and frequency analysis through Fourier transform.

\cite{Trabel18deepcom} provides building blocks for complex-valued neural networks. They present complex versions of linear layers, convolutional layers, batch normalization, initialization and different activation functions. They also comment on complex differentiability, referring to earlier works \cite{hirose2012gen}. 
Complex-valued building blocks have been used to develop a multitude of architectures, such as complex-valued generative adversarial networks \cite{vasudeva2022compressed, li2020sscv}, complex-valued convolutional recurrent networks \cite{hu2020dccrn} and a complex-valued U-net \cite{choi2018phase}. There has also been recent interest in optimizing computability on GPUs for complex-valued neural networks \cite{zhang2021optical}.

The transformer architecture \cite{vaswani2017attention} was a great success in natural language processing and has since become dominant in the field \cite{otter2020survey, hu2019introductory}. It has spread into vision \cite{Dosovitskiy2021an, khan2021transformers}, music \cite{Huan2019music} and more applications \cite{lin2021survey}. Additionally, there has been many works to improve and rework the architecture \cite{tay2020efficient}.

To the best of our knowledge, there are only two works concerning the design of a complex-valued transformer architecture.
\cite{yang2020complex} proposes a complex-valued transformer, motivated by the multiplicative structure of the Dot-Product attention.
They separate the product $Q(K^T)$ into eight real-valued products and then apply real-valued attention to all summands separately.
While being a well motivated choice, they use real-valued encoding matrices, making the network not fully complex-valued.
Using complex-valued encoding matrices would lead to a total of 64 summands and an unreasonable computational blowup.
While testing their framework against competitive models, such as real-valued transformers and LSTMs, they do not test against other definitions of complex-valued attention modules.
Additionally, in their experimental part, they do not use an independent test set but rather just evaluate on the validation set.
We answer the questions this work left open by incorporating their idea into our experiments and using it as one of many valid definitions for a complex-valued transformer.
\cite{dong2021signal} proposes a complex-valued meta-learning framework for signal recognition. As a byproduct, they define a complex-valued attention. However, they do not evaluate different options and they do not utilize the Dot-Product in the complex domain.
Additionally, they propose to use the complex variance for normalization instead of the more flexible covariance matrix. Their definition is, as one of many, incorporated in our framework.

There are some more works on different kinds of complex-valued attention modules~\cite{zhang2021end,ren2021polsar,cho2021complex}.
These use different kinds of convolutional architectures but are not complex-valued versions of the Scaled Dot-Product attention~\cite{vaswani2017attention}.

\section{Transformer}
This section serves as a brief description to the transformer architecture as introduced in \cite{vaswani2017attention}.

\subsection{Architecture}
\begin{figure}
	\begin{subfigure}{.3\linewidth}
		\scriptsize
		\begin{tcolorbox}[left = 0mm, right = 1mm, boxrule=0pt, colframe=white, colback=black!16, arc=0pt, outer arc=0pt]
		\textbf{Legend:}\vspace{1mm}\\
EE = Encoder \\ Embedding\vspace{1mm}\\
	DE = Decoder \\ Embedding\vspace{1mm}\\
	PE = Positional \\ Encoding\vspace{1mm}\\
(M)MHA = (Masked) Multi-Head Attention\vspace{1mm}\\
	LN = Layer \\ normalization\vspace{1mm}\\
	FF = Feed Forward\vspace{1mm}\\
	N x = repeat N times

	\end{tcolorbox}
	 \end{subfigure}
	 \begin{subfigure}{0.05\linewidth}
		 \hspace{0.05\linewidth}
	 \end{subfigure}
	\begin{subfigure}{.35\linewidth}
    \includegraphics[width=\linewidth]{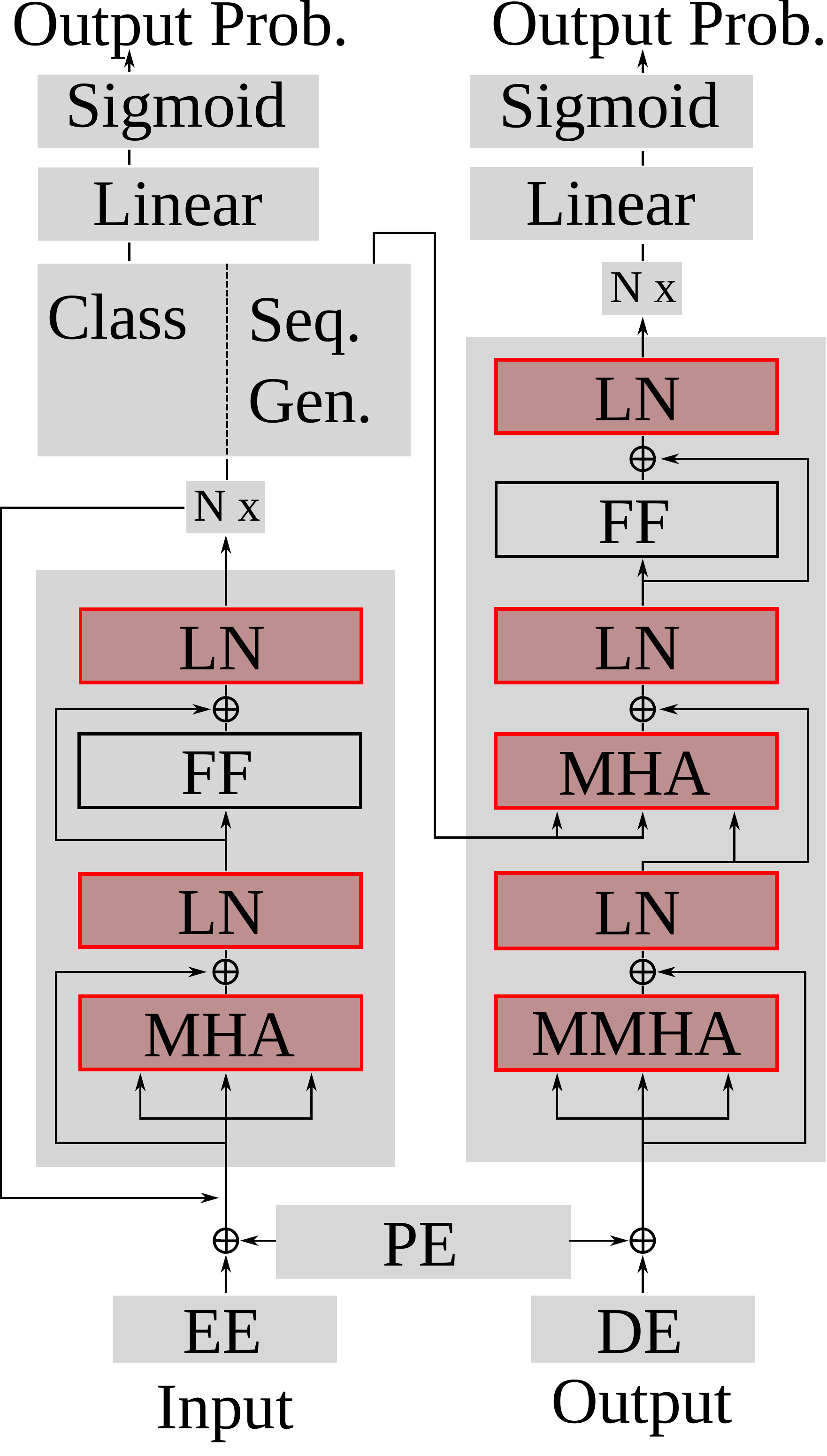}
	 \end{subfigure}
	 \begin{subfigure}{0.02\linewidth}
	 \hspace{0.02\linewidth}
	 \end{subfigure}
	 \vline
	 \begin{subfigure}{0.02\linewidth}
		 \hspace{0.02\linewidth}
	 \end{subfigure}
	\begin{subfigure}{.2\linewidth}
    \includegraphics[width=\linewidth]{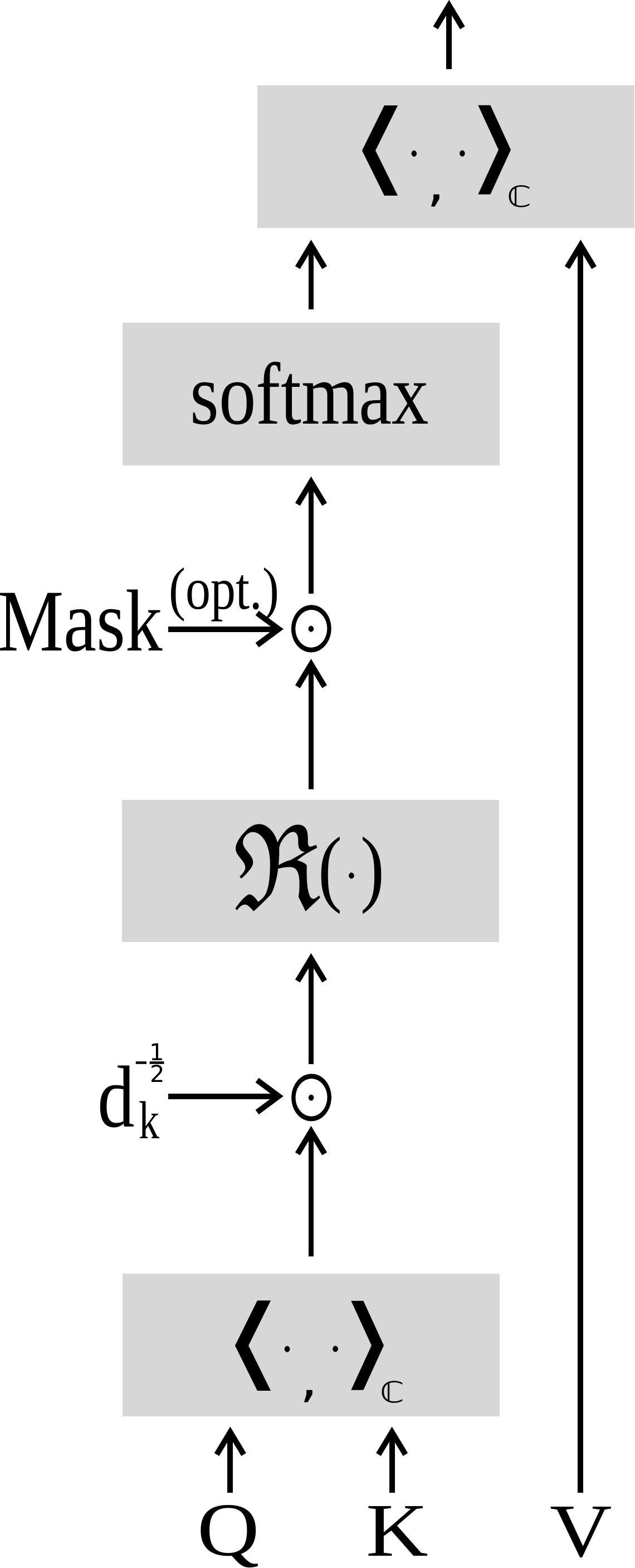}
	 \end{subfigure}
    \caption{Left: The general transformer architecture as introduced in \cite{vaswani2017attention}. Building blocks, whose complex-valued versions are derived in this paper are highlighted in red. Right: The complex-valued Scaled Dot-Product Attention as suggested in \autoref{subsec:c_att}.}
    \label{fig:architecture}
\end{figure}

The architecture consists of an encoder and decoder module.
The encoder module alone can be used for classification tasks while the full architecture can be used for sequence generation where the core idea is to input the original input into the encoder and the earlier outputs of the sequence generation into the decoder.
Both modules start with an embedding and a positional encoding of their respective inputs.
Afterwards the modules consist of Multi-Head Attention mechanism with residual connections followed by a layer normalization and a feed forward module - a small MLP with two linear layers and an activation. Details can be seen in \autoref{fig:architecture}.

\subsection{Real-valued Attention} \label{subsec:real_att}
The Scaled Dot-Product Attention is the core of the Transformer architecture.
The input consists of three matrices, called query $Q$, key $K$ and value $V$.
First the Dot-Product of the key and query is calculated, then scaled by the square-root of their equal dimensions $\sqrt{d_k}$.
The output is normalized by the softmax function and afterwards multiplied with the values.
Defining the softmax for a vector $X$ of length $n$ as
\begin{equation}
    softmax(X) = \sigma(X) = \frac{\exp(X)}{\sum_{i=1}^{n}\exp(X_i)}
\end{equation}
we can formulate the Scaled Dot-Product Attention as
\begin{equation}\label{eq:real_att}
    Att(Q, K, V) = \sigma\left(\frac{Q (K)^T}{\sqrt{d_k}}\right)V
\end{equation}
This core concept of Scaled Dot-Product Attention is extended to the more general concept of Multi-Head Attention (MHA) by applying learnable linear projections $W^Q$, $W^K$, $W^V$ to the inputs and project it back with another learnable linear projection $W^O$:
\begin{equation}
\begin{aligned}[b]\label{eq:mha}
    MHA(Q, K, V) = Concat(h_1, \hdots, h_k) W^O_i,\\
    \text{ where } h_i = Att(QW^Q_i, KW^K_i, VW^V_i)
\end{aligned}
\end{equation}
For the training process, it is necessary to mask future events in the decoder input. This is obtained by addition of $-\infty$ to the respective tokens before the softmax, which results in an attention score of $0$.

\section{Complex-valued Building Blocks}
The general purpose of this section is the introduction of existing building blocks and the development of new ones where needed (subsections \ref{subsec:c_att} and \ref{subsec:cln}) for a mathematically rigorous extension of the transformer architecture to the complex domain.
\cite{yang2020complex} has already introduced complex-valued fully connected feed forward layers.
Since the position within a sequence is real-valued, we adopt the sine and cosine positional encoding as originally used in \cite{vaswani2017attention} but other positional encodings would be possible \cite{wang2019encoding, chu2021conditional}.
\subsection{Complex-valued Attention}\label{subsec:c_att}
\begin{figure}[t]
    \centering
    \begin{tabular}{ccccc}
    \textcolor{orange}{Z}=$\langle \textcolor{myblue}{Q}, \textcolor{mygreen}{K} \rangle$ \hspace{-0.05\linewidth}
      & \includegraphics[width=.15\linewidth,valign=m]{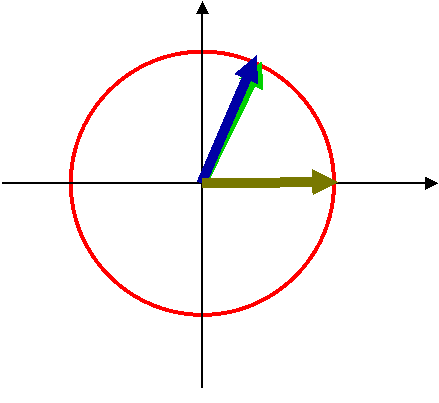}
      & \includegraphics[width=.15\linewidth,valign=m]{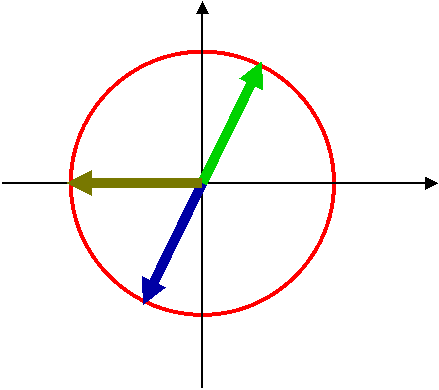}
      & \includegraphics[width=.15\linewidth,valign=m]{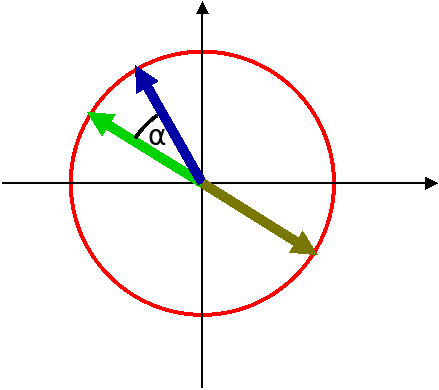}
      & \includegraphics[width=.15\linewidth,valign=m]{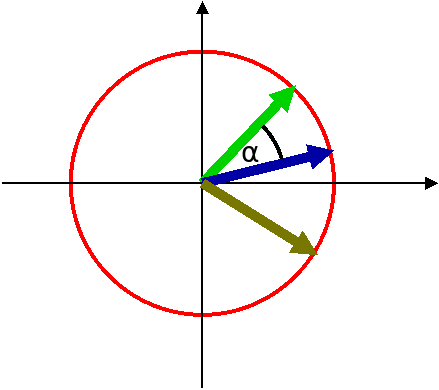}\\
      \textcolor{pink}{Z}=$\textcolor{myblue}{Q}(\textcolor{mygreen}{K}^T)$ \hspace{-0.05\linewidth}
      & \includegraphics[width=.15\linewidth,valign=m]{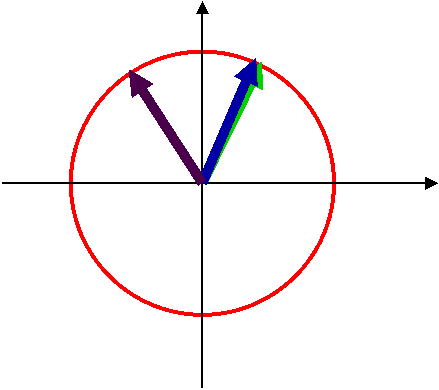}
      & \includegraphics[width=.15\linewidth,valign=m]{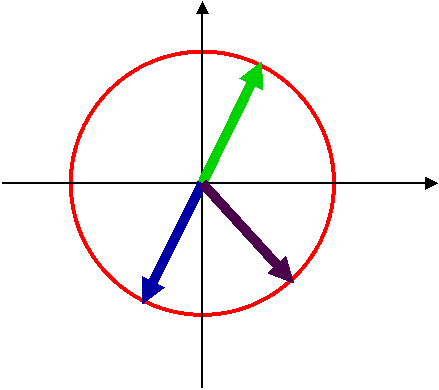}
      & \includegraphics[width=.15\linewidth,valign=m]{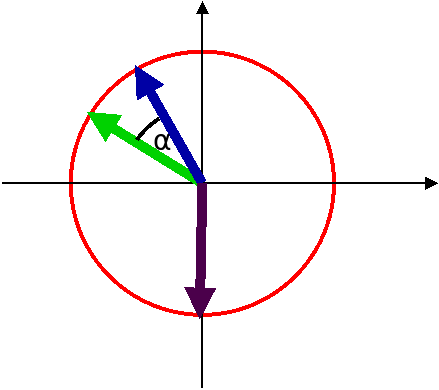}
      & \includegraphics[width=.15\linewidth,valign=m]{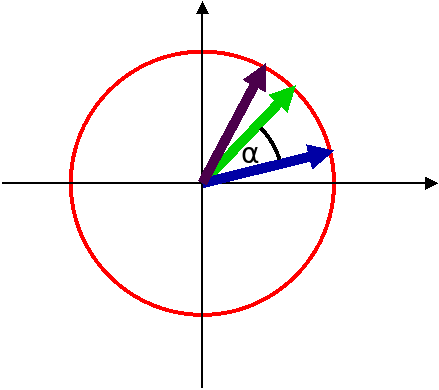}
    \end{tabular}
    \caption{ 
	    Illustration of Dot-Product vs $Q(K^T)$ in $\C$. 1st and 2nd column show behavior for the extreme cases of $Q=K$ and $Q=-K$. 3rd and 4th column show rotational invariance of the Dot-Product vs behavior on rotations of $Q(K^T)$. 
    }\label{fig:illustration_inner_prod}
\end{figure}

When generalizing the Scaled Dot-Product Attention to the complex domain a problem arises: the max operation does not work in $\C$ and the softmax does not either. However, the core idea behind the operation $\sigma(X Y^T)$ for $X, Y \in \R^n$ is to define a similarity between $X$ and $Y$ which is then scaled to $(0, 1)$ by the softmax non-linearity $\sigma$. The operation without softmax-rescaling can be described as the Dot-Product in $\R^n$. Using this concept in $\C^n$ leads to:
\begin{equation}\label{eq:dot_prod}
    \langle X, Y \rangle = \sum_{i=1}^n X_i \bar{Y}_i = \sum_{i=1}^n |X||Y|\exp(i(\phi_{X_i} - \phi_{Y_i}))
\end{equation}
When neglecting the magnitudes of $X$ and $Y$, we get
\begin{equation}\label{eq:euler}
    \exp(i(\phi_{X_i} - \phi_{Y_i})) = \cos(\phi_{X_i} - \phi_{Y_i}) + i \sin(\phi_{X_i} - \phi_{Y_i})
\end{equation}
The real part of this term, $\cos{\phi_{X_i} - \phi_{Y_i}}$, maximizes at $1$ for $\phi_{X_i} - \phi_{Y_i} = 0$, which is equivalent to $X_i=Y_i$. It strictly decreases, when $|\phi_{X_i} - \phi_{Y_i}|$ growth, up to a minimum of $-1$ at $|\phi_{X_i} - \phi_{Y_i}| = \pi$, which is equivalent to $X_i = -Y_i$.
Thus we have that $\real \left( \langle X, Y \rangle  \right)$ measures the similarity of $X$ and $Y$ for every component and adds these up.
Additionally, we get two desired properties for a similarity measure: Symmetry and rotational invariance.
Symmetry holds because the conjugate symmetry of the Dot-Product does not change its real part: 
\begin{equation}
	\real \langle Q, K \rangle = \real \overline{\langle K, Q \rangle} = \real \langle K, Q \rangle
\end{equation}
By rotational invariance we mean: If $Q$ and $K$ are (elementwise) both rotated by a fixed angle $\alpha$, $\langle Q, K \rangle$ does not change (\autoref{fig:illustration_inner_prod}).
\begin{align}
	\exp(i((\phi_{X_i} + \alpha)- (\phi_{Y_i} + \alpha))) = \exp(i(\phi_{X_i} - \phi_{Y_i}))
\end{align}
Note that the equality $\langle X, Y \rangle = Q(K^T)$ holds in the real domain, but not in the complex domain. While symmetry still holds when using $Q(K)^T$, the rotational invariance does not (\autoref{fig:illustration_inner_prod}).

When taking the magnitude into account, this similarity is scaled by the factors $|X|$ and $|Y|$, meaning that high values are obtained for vectors of high magnitude pointing in the same direction.
We can now formulate the complex-valued Dot-Product Attention as:
\begin{equation}\label{eq:rsoftmax}
    \C Att(A, B) = \sigma\left(\frac{\real\langle Q, K \rangle}{\sqrt{d_k}}\right) V
\end{equation}
This pipeline is presented on the right in \autoref{fig:architecture}.

The motivation of Scaled Dot-Product attention leads to \autoref{eq:rsoftmax}.
However, other possibilities to generalize the real-valued Scaled Dot-Product attention are using the absolute value with and without keeping the phase information and using both the real and the imaginary part.
We define the following possibilities and test all these in \autoref{sec:exp}. $|z|_\C$ denotes the absolute value of a complex number $z$ and $\sgn(z)$ its sign (e.g. $\frac{z}{|z|}$ if $z \neq 0$ and $1$ o/w):
\begin{align}
    AAtt(A,B) = & \sigma\left(\frac{\left|\langle Q, K \rangle \right|_\C}{\sqrt{d_k}}\right) V \\
    APAtt(A,B) = & \sigma\left(\frac{\left|\langle Q, K \rangle \right|_\C}{\sqrt{d_k}}\right) \sgn(\langle Q, K \rangle) V \\
    \real\imag Att(A,B) = & \left(\sigma\bigg( \frac{\real\langle Q, K \rangle}{\sqrt{d_k}}\right) + i \ \sigma\left(\frac{\imag\langle Q, K \rangle}{\sqrt{d_k}} \right)\bigg) V \label{eq:RIatt}
\end{align}
Additionally, it is possible to replace the dot product $\langle Q, K \rangle$ in every version with $Q \left(K \right)^T$. Using $Q \left(K \right)^T$, AAtt and $\C$Att have been used before \cite{dong2021signal}, we test these variants in \autoref{sec:exp}. Note $K\mapsto \overline{K}$ is not linear in $\C$ and thus cannot be learned directly by $W^K$.

Using any of these formulations of the complex-valued Scaled Dot-Product Attention the adoption of Multi-Head Attention as described in \autoref{subsec:real_att} is straightforward.
We can replace the learnable linear projections $W^Q$, $W^K$, $W^V$ and $W^O$ by complex-valued linear projections \cite{Trabel18deepcom} and can then use the formulations as described in \cite{vaswani2017attention} and \autoref{eq:mha}.
The necessary masking of future results in the training process as described in \autoref{subsec:real_att} works in this framework by applying the mask after the respective mappings from the complex to the real domain (such as $\real, \imag, |\cdot|$).

We also compare to the approach of \cite{yang2020complex} which relies on splitting the product $Q \left(K \right)^T$ into (real-valued) summands and applying real-valued attention per summand.

\subsection{Complex-valued Layer Normalization}\label{subsec:cln}
Normalization layers play a big role in the success of most neural network architectures.
A complex-valued version has been proposed for batch normalization \cite{Trabel18deepcom}, however layer normalization is preferable for methods like LSTM or RNNs~\cite{ba2016layer} as well as the transformer architecture~\cite{vaswani2017attention}. 
It is insufficient to normalize the real and imaginary part of the complex-valued layer independently, since this may lead to very elliptic shapes of the output distribution~\cite{Trabel18deepcom}.
This brings the need of a complex-valued version of layer normalization.
The necessary building blocks are the complex-valued expected value and the covariance matrix. For some complex vector $z\in \C^n$ these are defined as
\begin{align}
	\E(z) & = \frac{1}{n} \sum_{i=1}^n z_i \\
	\Cov(z) & = \begin{pmatrix} \Var(\real(z)) & \rCov(\real(z), \imag(z)) \\
	\rCov(\real(z), \imag(z)) & \Var(\imag(z))
	\end{pmatrix}
\end{align}
where $\Var$ and $\rCov$ denote the (real-valued) Variance and Covariance, respectively. 

Let $X$ be the output of a layer, the normalized output is then:
\begin{align}
	\begin{pmatrix}
		\real(\C LN(X)) \\
		\imag(\C LN(X))
	\end{pmatrix}
		= \Cov^{-\frac{1}{2}}(X) 	
	\begin{pmatrix}
		\real(X-\E(X)) \\
		\imag(X-\E(X))
	\end{pmatrix}
\end{align}
This compact form can easily be calculated with fast closed form solutions for the inverse and the square root of $2\times 2$ matrices.
It is possible to manipulate the output distribution with learnable parameters. It can be shifted with a learnable parameter $\beta \in \C$ and scaled with a learnable covariance matrix, a positive definite $2\times 2$ matrix. To ensure the positive definiteness of the resulting matrix, we utilize:
\begin{align}
	\begin{pmatrix}
		a & b \\
		b & c
	\end{pmatrix}
	\text{ positive definite} \Leftrightarrow a>0, c>0, b^2<ac
\end{align}
Thus, we can scale and shift the output $\hat{X}$ of the layer normalization with 5 degrees of freedom by learning a covariance matrix $\zeta$ and a shifting parameter $\beta\in\C$ and get:
\begin{align}
	\begin{pmatrix}
		\real(\hat{X}) \\
		\imag(\hat{X})
	\end{pmatrix}
	= \zeta^\frac{1}{2} \Cov^{-\frac{1}{2}}(X) 	
	\begin{pmatrix}
		\real(X-\E(X)) \\
		\imag(X-\E(X))
	\end{pmatrix} + \beta
\end{align}
The output distribution then has covariance $\zeta$ and expected value $\beta$.

\section{Experimental Results}\label{sec:exp}
Overall we perform two experiments: Automatic music transcription performed by the transformer encoder and a sequence generation task performed by the full transformer architecture.
We compare the introduced methods for a complex-valued attention module as described in Equations \ref{eq:rsoftmax}-\ref{eq:RIatt} using the proposed Dot-Product as well as the version using $Q(K^T)$.
Additionally, we compare to the approach of \cite{yang2020complex} and to the real-valued transformer as a baseline. For the latter, the real and imaginary part of the real-valued input was stacked alternating resulting in an input dimension of twice the original dimension.
Both tasks are trained and evaluated on the MusicNet dataset \cite{Thickstun2017learn}.
The dataset consists of 330 pieces of music divided into 39438 samples consisting of 64 time steps, which are interpreted as one input token.
These samples are split into 35111 training, 2030 validation and 3897 test samples, where the pieces of music between the splits do not overlap.
We perform Fourier transform on the data as preprocessing, as well as resampling as done in \cite{yang2020complex} with a method introduced by \cite{smith2002resample}.

For both experiments the important hyperparameters are: Batchsize 35, 100 epochs, dropout 0.1, learning rate $10^{-4}$, embedding dimension 320, 6 layers with 8 attention heads and a hidden dimension in the feed forward module of 2048.
As the encoder embedding we use a four layer complex-valued CNN followed by a fully connected layer. For the decoder embedding we used a fully connected embedding, since the input here are labels rather then a continuous signal.
To test the impact of the convolutions on the encoder, we perform a small ablation study on just the proposed method (\autoref{eq:rsoftmax}) by removing the CNN in the encoder embedding.

\begin{table}[t]
    \centering
    \begin{tabular}{c|c|c}
    Architecture & Classification & Seq. generation \\
\hline
    $\C$-Transformer (ours) & 14m & 27m\\

\hline
Yang et al \cite{yang2020complex} & 12m & 20m \\

\hline
$\R$-Transformer \cite{vaswani2017attention} & 18m & 33m  \\

\hline
    \end{tabular}
    \caption{Number of real-valued trainable parameters. For complex-valued parameters the real and imaginary parts count separately.}\label{tab:params}
\end{table}

\begin{table}[t]
    \centering
    \begin{tabular}{c|c|c||c|c}
 & \multicolumn{2}{c||}{Classification} & \multicolumn{2}{|c}{Sequence Generation} \\
	 \hline
Attention & Dot-Prod & $Q(K^T)$ & Dot-Prod & $Q(K^T)$ \\
\hline
$\C$Att		& 0.7164 & 0.7142 & 0.3272 & 0.3283 \\
\hline                                            
APAtt 		& 0.6965 & 0.6926 & 0.2240 & 0.3231 \\
\hline                                            
AAtt 		& 0.7117 & 0.7099 & 0.3172 & 0.3271 \\              
\hline                              
$\real\imag$Att & 0.7070 & 0.7059 & 0.3201 & 0.3236 \\
\hline
Yang et al \cite{yang2020complex} &x& 0.7088 & x      & 0.3072 \\
\hline
Real \cite{vaswani2017attention} & 0.7109 & x & 0.0737 & x 
    \end{tabular}
    \caption{Average precision results on test set for both tasks. Dot-Prod refers to the use of Dot-Product as described in \autoref{subsec:c_att}.}\label{tab:test}
\end{table}

\begin{table}[t]
    \begin{tabular}{c|c|c}
 & Classification & Seq. generation \\
\hline
$\C$-Attention w/o conv. & 0.5240 & 0.1652 \\
\hline
$\C$-Attention with conv. & 0.7164 & 0.3272
        \end{tabular}
    \caption{Average precision results on test set for both task as a small ablation study on the impact of the convolutional encoder.}\label{tab:testabl}

\end{table}

\begin{figure}
    \begin{center}
        \includegraphics[width=0.468\linewidth]{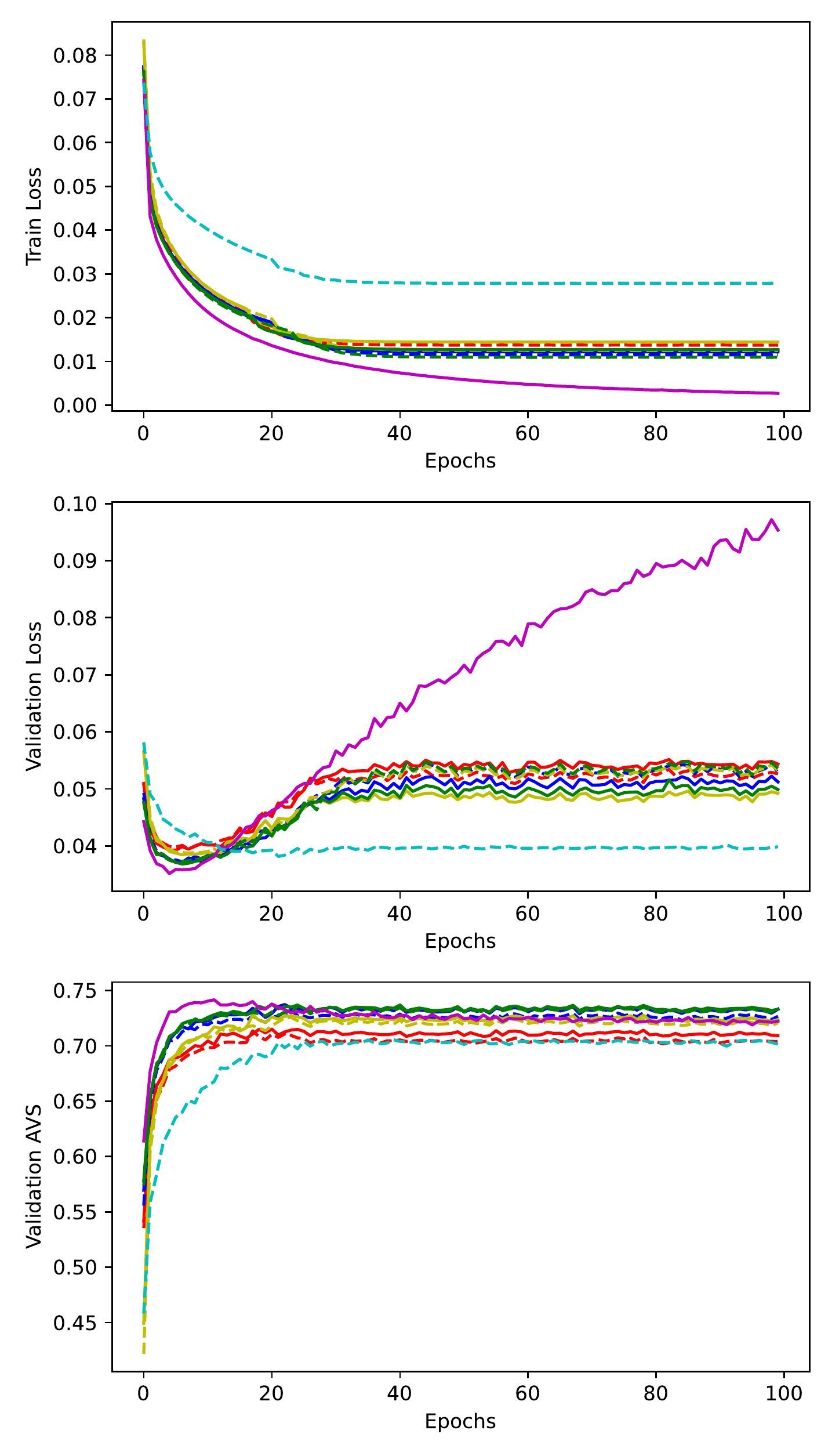}
        \includegraphics[width=0.47\linewidth]{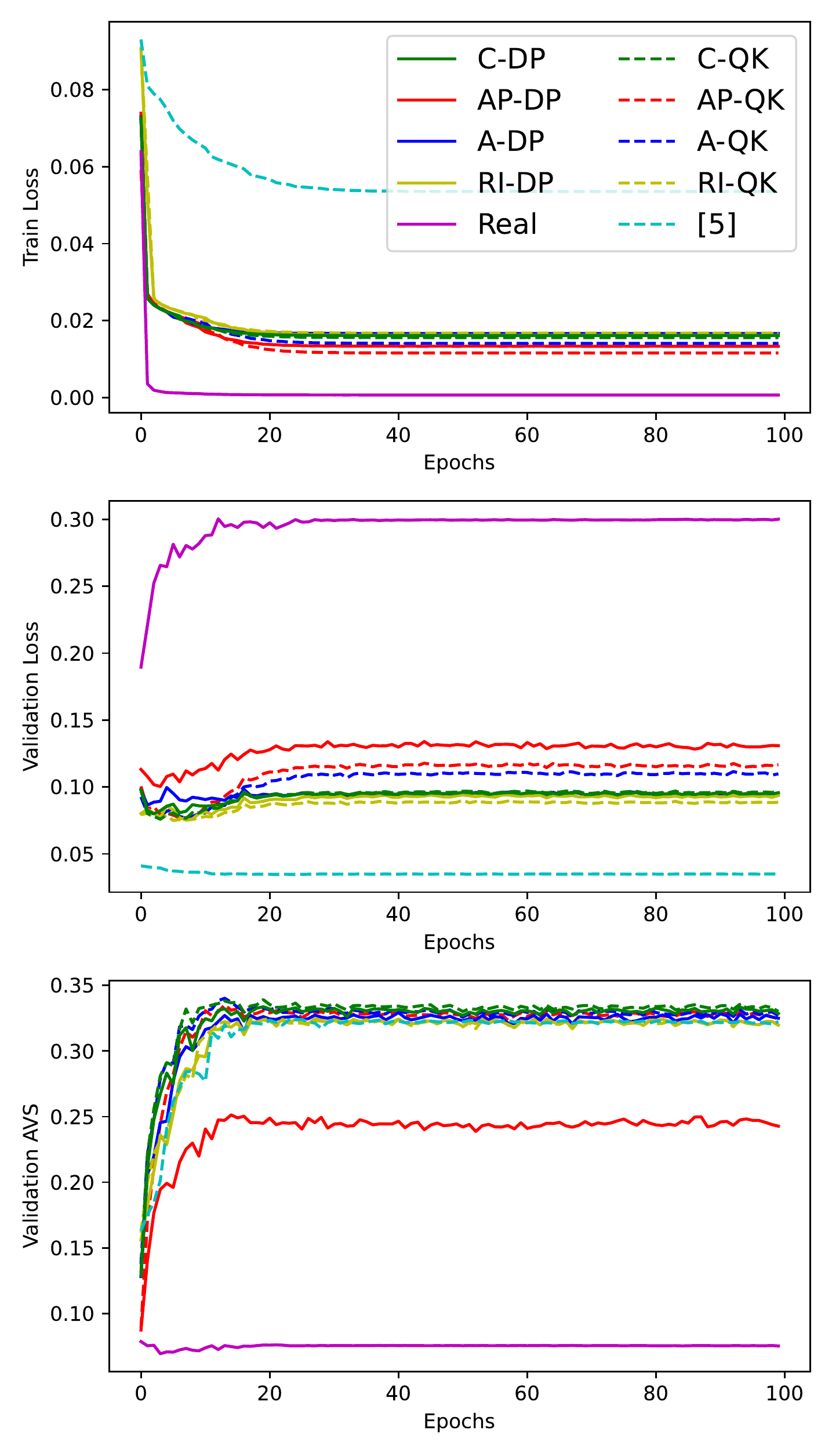}
\caption{Training and validation Results. Average Precision Score (AVS) and Loss are displayed. Left: Classification, right: Sequence generation. "DP" denotes Dot-Product, "QK" denotes $Q(K^T)$.}\label{fig:valres}
    \end{center}
\end{figure}

\subsection{Automatic music transcription task}
\label{subsec:class}
For the multiclass classification problem, we classify into 128 classes, as offered in the dataset.
The results in \autoref{tab:test} show competitive behavior of most methods after 100 epochs.
The best result by a slight margin is obtained by the proposed method as introduced in \autoref{eq:rsoftmax}.
For most attention variants, it shows that the inner product version performs better, than using the product $Q(K^T)$.
Additionally, all complex-valued architectures show improved robustness to overfitting (\autoref{fig:valres}), with no or minor decreases after longer training time, while the real transformer shows massive overfitting starting after 10 epochs.

\subsection{Sequence generation task}\label{subsec:gen}
For the sequence generation task, we split each sample into 43 input time steps and 21 time steps to be generated.
The output to be generated are the notes of the missing 21 time steps of the samples in an iterative way, where in each iteration the input of the decoder is the output of the earlier iterations.
The results show that the real transformer is not able to learn the sequence generation properly. The low training loss implies that the inability to learn is due to heavy overfitting. The robustness to overfitting already shown in \autoref{subsec:class} seems to solve this problem, where all complex-valued methods learn reasonably well. The best performance is again obtained by the proposed method \autoref{eq:rsoftmax}. While some of the other methods introduced in this paper perform similarly well, the method of \cite{yang2020complex} performs noticeably worse.

\subsection{Discussion}
\autoref{tab:testabl} shows that the complex-valued transformer architecture without convolutions is able to learn a meaningful solution, but the CNN in the encoder is necessary for state-of-the-art results.

Overall, the real transformer struggles with overfitting on both tasks.
The complex-valued transformers improve in this regard while maintaining on-par (\autoref{subsec:class}) or superior (\autoref{subsec:gen}) performance. This result is in line with earlier results, showing superior robustness to overfitting for, e.g., CNNs \cite{guberman2016complex} or RCNNs \cite{zhao2019semi}.
We are the first ones to show this for the transformer architecture.

\section{Conclusion}

We presented building blocks for a complex-valued transformer architecture. That includes newly developed formulations of complex-valued attention mechanisms as well as a complex-valued layer normalization.
We have shown that it improves robustness to overfitting on a classification and a sequence generation task, while maintaining competitive performance compared to the real-valued algorithm when applied to complex-valued signals.
This opens up the opportunity to incorporate the transformer architecture into a more broad class of applications that naturally make use of complex-valued images. 
Additionally, the complex-valued Fourier transform of signals can now directly be used in the transformer architecture without using the isomorphism $\C \to \R^2$ that results in a loss in robustness against overfitting.
This work also serves as a base for the further development of complex-valued versions of extensions to the transformer \cite{tay2020efficient} and other architectures that use the attention mechanism.

\vfill
\pagebreak


\bibliographystyle{IEEbib}
\bibliography{bib}

\begin{thebibliography}{10}

\bibitem{cole2021analysis}
E.~Cole, J.~Cheng, J.~Pauly, and S.~Vasanawala,
\newblock ``Analysis of deep complex-valued convolutional neural networks for
  {MRI} reconstruction and phase-focused applications,''
\newblock {\em Magnetic Resonance in Medicine}, vol. 86, no. 2, pp. 1093--1109,
  2021.

\bibitem{vasudeva2022compressed}
B.~Vasudeva, P.~Deora, S.~Bhattacharya, and P.~M. Pradhan,
\newblock ``Compressed sensing mri reconstruction with {Co-VeGAN}:
  Complex-valued generative adversarial network,''
\newblock in {\em WACV}, 2022, pp. 1779--1788.

\bibitem{li2020sscv}
X.~Li, Q.~Sun, L.~Li, X.~Liu, H.~Liu, L.~Jiao, and F.~Liu,
\newblock ``{SSCV-GANs}: Semi-supervised complex-valued {GANs} for {PolSAR}
  image classification,''
\newblock {\em IEEE Access}, vol. 8, pp. 146560--146576, 2020.

\bibitem{choi2018phase}
H.-S. Choi, J.-H. Kim, J.~Huh, A.~Kim, J.-W. Ha, and K.~Lee,
\newblock ``Phase-aware speech enhancement with deep complex {U-Net},''
\newblock in {\em ICLR}, 2018.

\bibitem{yang2020complex}
M.~Yang, M.~Q. Ma, D.~Li, Y.-H.~H. Tsai, and R.~Salakhutdinov,
\newblock ``Complex transformer: A framework for modeling complex-valued
  sequence,''
\newblock in {\em ICASSP}, 2020, pp. 4232--4236.

\bibitem{yang2020fda}
Y.~Yang and S.~Soatto,
\newblock ``{FDA}: Fourier domain adaptation for semantic segmentation,''
\newblock in {\em CVPR}, 2020, pp. 4084--4094.

\bibitem{xu2021fourier}
Q.~Xu, R.~Zhang, Y.~Zhang, Y.~Wang, and Q.~Tian,
\newblock ``A {Fourier}-based framework for domain generalization,''
\newblock in {\em CVPR}, 2021, pp. 14383--14392.

\bibitem{yin2020phasen}
D.~Yin, C.~Luo, Z.~Xiong, and W.~Zeng,
\newblock ``{PHASEN}: A phase-and-harmonics-aware speech enhancement network,''
\newblock in {\em AAAI}, 2020.

\bibitem{Trabel18deepcom}
C.~Trabelsi, O.~Bilaniuk, Y.~Zhang, D.~Serdyuk, S.~Subramanian, J.~Felipe
  Santos, S.~Mehri, N.~Rostamzadeh, Y.~Bengio, and C.~J. Pal,
\newblock ``Deep complex networks,''
\newblock in {\em ICLR}, 2018.

\bibitem{vaswani2017attention}
A.~Vaswani, N.~Shazeer, N.~Parmar, J.~Uszkoreit, L.~Jones, A.~N. Gomez,
  L.~Kaiser, and I.~Polosukhin,
\newblock ``Attention is all you need,''
\newblock in {\em NIPS}, 2017, pp. 5998--6008.

\bibitem{kataoka1998music}
M.~Kataoka, M.~Kinouchi, and M.~Hagiwara,
\newblock ``Music information retrieval system using complex-valued recurrent
  neural networks,''
\newblock in {\em IEEE SMC}, 1998, pp. 4290--4295.

\bibitem{kuroe2002energy}
Y.~Kuroe, N.~Hashimoto, and T~Mori,
\newblock ``On energy function for complex-valued neural networks and its
  applications,''
\newblock in {\em ICONIP}, 2002.

\bibitem{hirose2003complex}
A.~Hirose,
\newblock {\em Complex-Valued Neural Networks: Theories and Applications},
\newblock World Scientific, 2003.

\bibitem{bassey2021survey}
J.~Bassey, L.~Qian, and X.~Li,
\newblock ``A survey of complex-valued neural networks,''
\newblock {\em arXiv:2101.12249}, 2021.

\bibitem{zhang2021end}
Y.-P. Zhang, Q.~Zhang, Le~Kang, Y.~Luo, and L.~Zhang,
\newblock ``End-to-end recognition of similar space cone--cylinder targets
  based on complex-valued coordinate attention networks,''
\newblock {\em IEEE Transactions on Geoscience and Remote Sensing}, vol. 60,
  pp. 1--14, 2021.

\bibitem{ren2021polsar}
S.~Ren and F.~Zhou,
\newblock ``Polsar image classification with complex-valued residual attention
  enhanced {U-Net},''
\newblock in {\em IGARSS}, 2021, pp. 3045--3048.

\bibitem{virtue2017better}
P.~Virtue, X.~Yu Stella, and M.~Lustig,
\newblock ``Better than real: Complex-valued neural nets for {MRI}
  fingerprinting,''
\newblock in {\em ICIP}, 2017, pp. 3953--3957.

\bibitem{hirose2012gen}
A.~Hirose and S.~Yoshida,
\newblock ``Generalization characteristics of complex-valued feedforward neural
  networks in relation to signal coherence,''
\newblock {\em IEEE Transactions on Neural Networks and Learning Systems}, vol.
  23, no. 4, pp. 541--551, 2012.

\bibitem{hu2020dccrn}
Y.~Hu, Y.~Liu, S.~Lv, M.~Xing, S.~Zhang, Y.~Fu, J.~Wu, B.~Zhang, and L.~Xie,
\newblock ``{DCCRN}: Deep complex convolution recurrent network for phase-aware
  speech enhancement,''
\newblock in {\em INTERSPEECH}, 2020.

\bibitem{zhang2021optical}
H.~Zhang et~al.,
\newblock ``An optical neural chip for implementing complex-valued neural
  network,''
\newblock {\em Nature Communications}, vol. 12, pp. 457, 2021.

\bibitem{otter2020survey}
D.~W. Otter, J.~R. Medina, and J.~K. Kalita,
\newblock ``A survey of the usages of deep learning for natural language
  processing,''
\newblock {\em IEEE Transactions on Neural Networks and Learning Systems}, vol.
  32, no. 2, pp. 604--624, 2020.

\bibitem{hu2019introductory}
D.~Hu,
\newblock ``An introductory survey on attention mechanisms in {NLP} problems,''
\newblock in {\em IntelliSys}, 2019.

\bibitem{Dosovitskiy2021an}
A.~Dosovitskiy, L.~Beyer, A.~Kolesnikov, D.~Weissenborn, X.~Zhai,
  T.~Unterthiner, M.~Dehghani, M.~Minderer, G.~Heigold, S.~Gelly, J.~Uszkoreit,
  and N.~Houlsby,
\newblock ``An image is worth 16x16 words: Transformers for image recognition
  at scale,''
\newblock in {\em ICLR}, 2021.

\bibitem{khan2021transformers}
S.~Khan, M.~Naseer, M.~Hayat, S.~W. Zamir, F.~S. Khan, and M.~Shah,
\newblock ``Transformers in vision: A survey,''
\newblock {\em ACM Computing Surveys}, vol. 54, pp. 200:1--200:41, 2021.

\bibitem{Huan2019music}
C.-Z.~A. Huang, A.~Vaswani, J.~Uszkoreit, I.~Simon, C.~Hawthorne, N.~Shazeer,
  A.~M. Dai, M.~D. Hoffman, M.~Dinculescu, and D.~Eck,
\newblock ``Music transformer: Generating music with long-term structure,''
\newblock in {\em ICLR}, 2019.

\bibitem{lin2021survey}
T.~Lin, Y.~Wang, X.~Liu, and X.~Qiu,
\newblock ``A survey of transformers,''
\newblock {\em AI Open}, vol. 3, pp. 111--132, 2022.

\bibitem{tay2020efficient}
Y.~Tay, M.~Dehghani, D.~Bahri, and D.~Metzler,
\newblock ``Efficient transformers: A survey,''
\newblock {\em ACM Computing Surveys}, vol. 55, no. 6, pp. 109:1--109:28, 2023.

\bibitem{dong2021signal}
Y.~Dong, Y.~Peng, M.~Yang, S.~Lu, and Q.~Shi,
\newblock ``Signal transformer: Complex-valued attention and meta-learning for
  signal recognition,''
\newblock {\em arXiv:2106.04392}, 2021.

\bibitem{cho2021complex}
H.-W. Cho, S.~Choi, Y.-R. Cho, and J.~Kim,
\newblock ``Complex-valued channel attention and application in ego-velocity
  estimation with automotive radar,''
\newblock {\em IEEE Access}, vol. 9, pp. 17717--17727, 2021.

\bibitem{wang2019encoding}
B.~Wang, D.~Zhao, C.~Lioma, Q.~Li, P.~Zhang, and J.~G. Simonsen,
\newblock ``Encoding word order in complex embeddings,''
\newblock in {\em ICLR}, 2020.

\bibitem{chu2021conditional}
X.~Chu, Z.~Tian, B.~Zhang, X.~Wang, X.~Wei, H.~Xia, and C.~Shen,
\newblock ``Conditional positional encodings for vision transformers,''
\newblock {\em arXiv:2102.10882}, 2021.

\bibitem{ba2016layer}
J.~L. Ba, J.~R. Kiros, and G.~E. Hinton,
\newblock ``Layer normalization,''
\newblock {\em arXiv:1607.06450}, 2016.

\bibitem{Thickstun2017learn}
J.~Thickstun, Z.~Harchaoui, and S.~M. Kakade,
\newblock ``Learning features of music from scratch,''
\newblock in {\em ICLR}, 2017.

\bibitem{smith2002resample}
J.~O. Smith,
\newblock ``Digital audio resampling home page,''
  \url{https://ccrma.stanford.edu/~jos/resample/}", 2020.

\bibitem{guberman2016complex}
N.~Guberman,
\newblock ``On complex valued convolutional neural networks,''
\newblock {\em arXiv:1602.09046}, 2016.

\bibitem{zhao2019semi}
F.~Zhao, G.~Ma, W.~Xie, and H.~Liu,
\newblock ``Semi-supervised recurrent complex-valued convolution neural network
  for polsar image classification,''
\newblock in {\em IGARSS}, 2019.

\end{thebibliography}

\end{document}